\documentclass[runningheads]{llncs}
\usepackage{graphicx}
\usepackage{cite}
\usepackage{amsmath,amssymb,amsfonts}
\usepackage{algorithmic}
\usepackage{graphicx}
\usepackage{textcomp}
\usepackage{adjustbox}
\usepackage{multirow}
\usepackage{bbold}
\usepackage{siunitx}
\usepackage{hyperref}       
\usepackage{xcolor}
\usepackage{sidecap}

\usepackage{booktabs}
\usepackage[mathscr]{eucal}
\usepackage{marvosym}

\hypersetup{
    colorlinks,
    linkcolor={blue!50!black},
    citecolor={blue!50!black},
    urlcolor={blue!50!black}
}
\usepackage{amsmath, bm}
\usepackage{hyperref}
\usepackage[title]{appendix}

\usepackage{pifont}

\usepackage{floatrow}
\newfloatcommand{capbtabbox}{table}[][\FBwidth]

\begin{document}

\title{
Benchmarking Robustness of Endoscopic Depth Estimation with Synthetically Corrupted Data
}
\titlerunning{Benchmarking Robustness for Depth Perception in Endoscopic Surgery}
\author{An Wang\inst{1} \and
Haochen Yin\inst{1} \and
Beilei Cui\inst{1} \and
Mengya Xu\inst{1} \and
Hongliang Ren\inst{\textsuperscript{\Letter},1,2}}

\authorrunning{Wang et al.}
%

\institute{Dept. of Electronic Engineering, The Chinese University of Hong Kong, Hong Kong SAR, China 
\and Dept. of Biomedical Engineering, National University of Singapore, Singapore \\
\email{hlren@ee.cuhk.edu.hk}}

\maketitle              
\begin{abstract}

Accurate depth perception is crucial for patient outcomes in endoscopic surgery, yet it is compromised by image distortions common in surgical settings. To tackle this issue, our study presents a benchmark for assessing the robustness of endoscopic depth estimation models. We have compiled a comprehensive dataset that reflects real-world conditions, incorporating a range of synthetically induced corruptions at varying severity levels.
To further this effort, we introduce the Depth Estimation Robustness Score (\textit{DERS}), a novel metric that combines measures of error, accuracy, and robustness to meet the multifaceted requirements of surgical applications. This metric acts as a foundational element for evaluating performance, establishing a new paradigm for the comparative analysis of depth estimation technologies.
Additionally, we set forth a benchmark focused on robustness for the evaluation of depth estimation in endoscopic surgery, with the aim of driving progress in model refinement. A thorough analysis of two monocular depth estimation models using our framework reveals crucial information about their reliability under adverse conditions.
Our results emphasize the essential need for algorithms that can tolerate data corruption, thereby advancing discussions on improving model robustness. The impact of this research transcends theoretical frameworks, providing concrete gains in surgical precision and patient safety. This study establishes a benchmark for the robustness of depth estimation and serves as a foundation for developing more resilient surgical support technologies.
Code is available at~\url{https://github.com/lofrienger/EndoDepthBenchmark}.


\end{abstract}

\section{Introduction}
The evolution of endoscopic surgery has led to significant advancements in the minimally invasive treatment of patients, offering reduced recovery times and smaller incisions compared to traditional surgery. However, a prominent challenge in the field is the surgeon's ability to accurately perceive depth through the two-dimensional video feed provided by the endoscope~\cite{bogdanova2016depth}. This depth perception is crucial for accurately manipulating instruments within the body, directly affecting the outcome of these intricate procedures. 

Technological advancements are compensating for endoscopic imagery's limited depth cues. 3D endoscopic systems, while accurate, are often sidelined due to cost and complexity~\cite{tomazic20213d}. Monocular systems prevail since they are affordable and easily integrated into surgical setups, despite depth perception limitations~\cite{breedveld1999theoretical}. Computational depth estimation models are thus crucial for assisting surgeons using monocular setups, providing depth information from 2D images~\cite{wang2017autostereoscopic}. Recent progress in this field leans towards self-supervised learning, which circumvents the need for actual depth data that is often hard to acquire in medical contexts~\cite{liu2018self,shao2022self,yang2024self}. These models use video sequence frame consistency to infer depth, employing temporal information~\cite{liu2019dense,widya2021self,shao2021self}.

\begin{figure}[t]
    \centering
    \includegraphics[width=\linewidth]{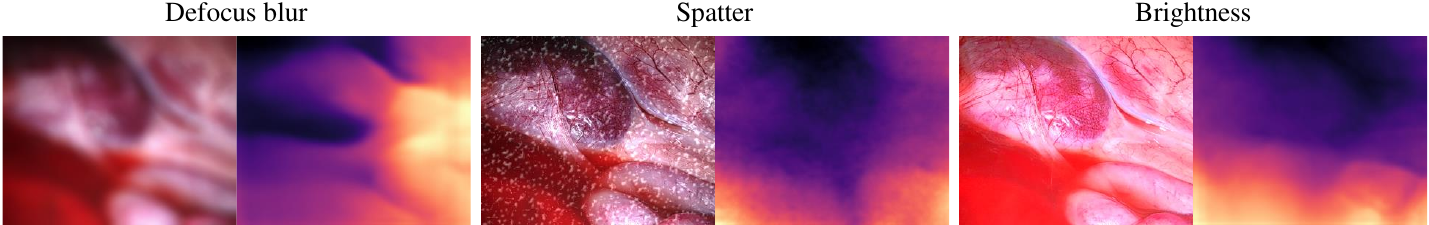}
    \caption{Inconsistent depth prediction under various data corruptions.}
    \label{fig:corr_depth}
\end{figure}

Despite noteworthy advancements, the effectiveness of such models may be compromised under real-world distortions like poor illumination and visual obstructions, thus impacting surgical outcomes~\cite{hofmeister2001perceptual}, as shown in Fig.~\ref{fig:corr_depth}. Consequently, the design of benchmarks assessing the robustness against image corruption has become crucial. While traditional benchmarks focus on precision under ideal conditions, studies aiming for robustness incorporate adverse scenarios, modeling noise, blur, occlusion, and exposure changes~\cite{xian2024towards,kong2023robo3d,mancini2016fast}. Nonetheless, these generalized corruption studies may not adequately address the unique challenges in endoscopic depth estimation, highlighting the need for specialized robustness benchmarks.

To address this limitation and dissect model stability, we introduce the robustness-centric benchmark by synthesizing a collection of varied types of corrupted data and propose the Depth Estimation Robustness Score (\textit{DERS}), a novel metric specifically designed to evaluate and contrast the performance of monocular depth estimation models within a controlled setting that replicates visual imperfections encountered in surgeries.
This benchmarking tool aims to assess not only a model's accuracy in ideal conditions but also its robustness against various types of image degradation, thus providing a more realistic assessment of a model's practical utility.

\textit{DERS} sets a precedent in performance evaluation by introducing a composite scoring system that integrates measures of error, accuracy, and robustness. With this metric, we aim to capture the multifaceted challenges of endoscopic surgery, pushing the development of depth estimation solutions that are not only precise but also reliable under less-than-ideal conditions~\cite{ozyoruk2021endoslam}.

To sum up, our contributions are as follows:
\begin{itemize}
    \item We examine real-world image degradation in endoscopic imaging and generate a comprehensive dataset consisting of various types of corrupted images with different degrees of severity.
    \item We propose the Depth Estimation Robustness Score (\textit{DERS}), which establishes a new standard for performance comparison of depth estimation models by introducing a composite scoring system that incorporates metrics of error, accuracy, and robustness.
    \item We develop a robustness-centric benchmark and toolkit for assessing depth perception in endoscopic surgery, supporting advances in resilient depth estimation models to improve surgical outcomes.
    \item By thoroughly evaluating two endoscopic monocular depth estimation models using our protocol, we engage in a detailed discussion on training corruption-tolerant endoscopic depth estimation models.
\end{itemize}

\section{Method}

\subsection{Common Endoscopic Corruptions}
\label{sec:corruptions}

\begin{figure}[!t]
    \centering
    \includegraphics[width=\linewidth]{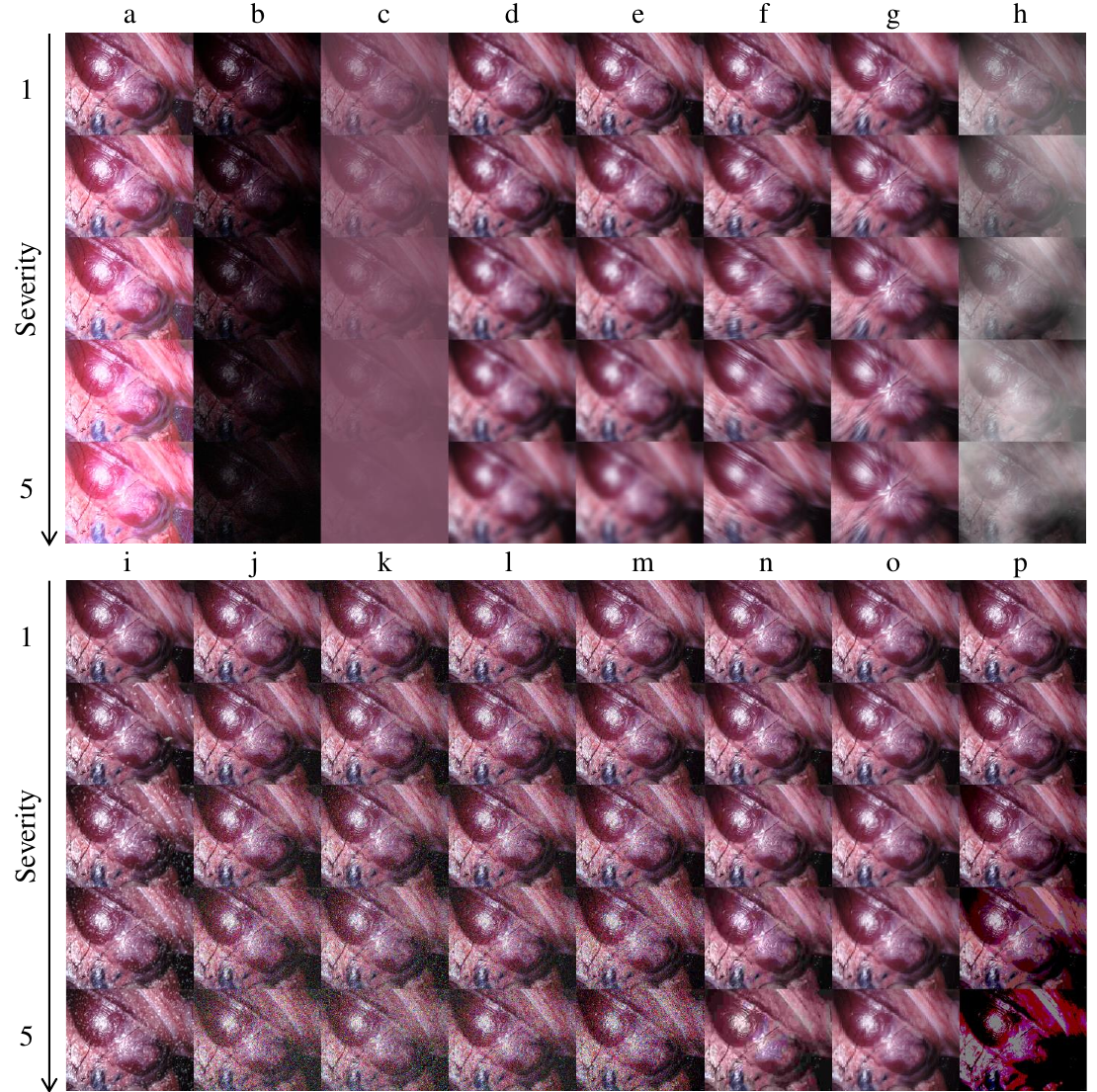}
    \caption{Depiction of the impact of various corruptions across five levels of severity. Columns a to p correspond to the respective corruption type introduced in Sec.~\ref{sec:corruptions}. Zoom in for the best views.}
    \label{fig:comp_corr}
\end{figure}

Endoscopic image quality is crucial for accurate navigation and decision-making in minimally invasive procedures. Referring to previous works~\cite{hendrycks2019robustness,kong2023robo3d,wang2023curriculum}, as shown in Fig.~\ref{fig:comp_corr}, we identify the primary corruption categories that affect endoscopic vision, each of which presents distinct types.

\textbf{Illumination Variability} (\textit{Brightness}, \textit{Dark}, \textit{Contrast})
Inconsistent lighting within endoscopic scenes can lead to overexposure or underexposure, both of which hide crucial details and hinder algorithms' interpretative capabilities. Abrupt contrast levels further complicate image analysis by distorting visible structures and potentially flattening depth perception.

\textbf{Optic Distortions} (\textit{Defocus Blur}, \textit{Motion Blur}, \textit{Zoom Blur}, \textit{Gaussian Blur})
Constraints and motions of camera optics give rise to a variety of blurs—from defocus to motion-induced smearing and zoom-related radial distortions—that reduce image clarity. Even non-optical, generalized Gaussian blur introduces a universal softening of detail that, while not sourced from camera movements, still simulates focus loss.

\textbf{Visual Obstructions} (\textit{Smoke}, \textit{Spatter})
For endoscopic procedures, visual obstructions such as smoke from cauterization and spatter from fluids or debris directly impact the surgical field's visibility. These specialized corruptions are particularly disruptive as they may directly cover the lens, immediately distorting and blocking vital visual information.

\textbf{Sensor and Electronic Noise} (\textit{Gaussian Noise}, \textit{Impulse Noise}, \textit{Shot Noise}, \textit{ISO Noise})
Introduced by the sensor and electronic elements, especially in low-light scenarios, a variety of noise artifacts can randomly alter pixel intensities or create scattered specks across the imagery. Attempts to amplify the signal through higher ISO settings inadvertently exacerbate noise, frequently leading to a more granulated image output.

\textbf{Data Compression and Digital Artifacts} (\textit{JPEG Compression}, \textit{Pixelated}, \textit{Color Quantization})
These problems are further exacerbated by artifacts that arise from necessary image compression. Whether it is lossy methods such as JPEG, which generates artifacts around areas of high contrast, or pixelation and color banding that result from aggressive compression or color quantization, these digital artifacts can lead to a reduction in image resolution. Consequently, this can significantly impede the identification of crucial anatomical features.

\subsection{Comprehensive Evaluation Metrics}

In assessing the performance of monocular depth estimation models, particularly within the challenging context of endoscopic surgery, a comprehensive evaluation framework is essential.

\noindent \textbf{Error-Accuracy Metrics}
Most widely used metrics in monocular depth estimation are error metrics, including Absolute Relative Difference ($AbsRel$), Squared Relative Difference ($SqRel$), Root Mean Squared Error ($RMSE$) and Root Mean Squared Error in Logarithmic Scale ($LogRMSE$), and accuracy metrics like Thresholded Accuracy ($a1$, $a2$, and $a3$). These metrics assess how accurately a model predicts depth by comparing its output to actual depth measurements. $AbsRel$ and $SqRel$ measure the average error in prediction relative to the true depth, $RMSE$ quantifies the overall error magnitude, while $a1$, $a2$, and $a3$ gauge the percentage of predictions within specific error margins. Together, they offer a comprehensive view of model performance, crucial for applications requiring precise depth information.

\subsubsection{Depth Estimation Robustness Score}
To measure a model's performance across various distortions common in endoscopic imaging, we present the Depth Estimation Robustness Score (\textit{DERS}). \textit{DERS} purposefully devised to combine three pivotal components—error, accuracy, and robustness—into a comprehensive composite index. Below, we examine the individual components of \textit{DERS} and explain how they are integrated within the structure of the metric.

(1) \textit{Error Component} $E$ 

The error component integrates four aforementioned traditional depth estimation metrics that methodically assess the variance between predicted and actual depth. For each metric, the error is normalized relative to the error from an uncorrupted (clean) image. This normalization process stabilizes the error values, thereby setting the clean image as a standard for comparison.
Specifically, we can formulate the error component as
\begin{equation}
\label{eq:err}
    E=\sum_{i=1}^4\frac{\sum_{j=1}^{m}E_{ij}}{m \cdot E_{i0}},
\end{equation}
where $E_{ij}$ is the error for the $i$-th metric at the corruption level $j$. $m=5$ is the total number of corruption levels. Note that we only average over the error metric on corrupted data.
$E_{i0}$ is the error metric $i$ for the clean (uncorrupted) image, which acts as the normalization factor.

(2) \textit{Accuracy Component} $A$

Accuracy delves into the proximity of the predicted depth values falling within certain thresholds relative to the true depths, denoted as $a1$, $a2$, and $a3$. These thresholds are defined by predetermined factors $\delta$ ($<1.25$, $<1.25^2$, $<1.25^3$), evaluating the proportion of accurate predictions amidst varying degrees of stringency. Each accuracy metric is computed and then weighted according to its significance. Concretely, the accuracy component can be expressed as
\begin{equation}
\label{eq:acc}
A=\sum_{k=1}^3 \frac{W_k}{m+1}\sum_{j=0}^{m}A_{kj},
\end{equation}
where $A_{kj}$ denotes the accuracy of the $k$-th metric for the corruption level $j$. We compute the average across all sets of data, capturing the general accuracy of the model regardless of the corruption level applied. $W_k$ is the weight assigned to the $k$-th accuracy threshold. These weights reflect the relative importance of different accuracy levels and are typically set to emphasize more stringent thresholds.

(3) \textit{Robustness Component} $R$

Robustness reflects the model's resilience to corruption, calculated as the standard deviation of all $7$ metric values across various levels of corruption relative to the performance on the clean image. This provides an empirical assessment of the model's consistency in performance and can be formulated as:

\begin{equation}
\label{eq:rob}
    R=\dfrac{\lambda}{7}\sum_{i=1}^{7}\sqrt{\dfrac{1}{m}\sum_{j=1}^{m}(M_{ij}-M_{i0})^2}.
\end{equation}

Here, $M_{ij}$ is the value for metric $i$ at corruption level $j$, and $M_{i0}$ is the metric value from the clean image. The robustness factor $\lambda$ acts as a tunable parameter, signifying the relative influence of model stability in the face of corruption when compared to the base accuracy and error performance. A higher robustness factor would put more emphasis on penalizing fluctuations in performance across levels of corruption, whereas a lower value would reduce its influence, potentially favoring models with better accuracy and error rates, despite possibly less consistency across varying conditions.
This formulation ensures that robustness reflects the model's performance variability relative to the baseline metrics of clean images. A lower value of $R$ signifies enhanced robustness in the model's performance.

(4) \textit{Final Hybrid Formulation} \textit{DERS}

Rendered as a whole, the Depth Estimation Robustness Score (\textit{DERS}) integrates the aforementioned components, thus:

\begin{equation}
    DERS=\frac{E}{A}\times e^{-R}.
\end{equation}

For clarity, it is anticipated that lower values of $E$ and $R$, combined with a higher value of $A$, will yield increased accuracy and robustness, thereby contributing to a superior overall robustness score.
Furthermore, incorporating error normalization, accuracy weighting, and performance degradation scaling into a singular measure, the \textit{DERS} offers a broad assessment of a depth estimation model's resilience against various endoscopic image corruptions. This metric delivers a more comprehensive and nuanced understanding of the model's effectiveness, giving it a competitive advantage over existing metrics that only assess error or accuracy.

\section{Experimental Settings}

\subsubsection{Dataset Generation}

Our study utilizes the SCARED~\cite{allan2021stereo} dataset, a collection unveiled during the MICCAI 2019 grand challenge\footnote{\url{https://endovissub2019-scared.grand-challenge.org}}, featuring 35 endoscopic videos equating to 22,950 frames. These frames offer a detailed display of abdominal anatomy within fresh porcine cadavers, all captured through the advanced optics of the da Vinci Xi endoscope. 
We segmented the dataset following the method proposed by Shao et al.~\cite{shao2022self}, allocating 551 frames for the test set.
Building upon SCARED~\cite{allan2021stereo}, we present an augmentation of this dataset, which we have termed SCARED-C. SCARED-C derives from the original but incorporates 16 variations of environmental and artificial corruptions, systematically applied at five different severity levels. 
We employ the severity settings from the imagecorruptions library\footnote{\url{https://github.com/bethgelab/imagecorruptions}}, with a minor modification to reduce the intensity of the ``Spatter'' corruption. Notably, a severity level of 0 signifies that the original image remains uncorrupted, while levels 1 through 5 represent progressive intensities of image distortions.
This expanded dataset serves as a rigorous evaluation platform for the accuracy of depth estimation within endoscopic imagery, hence serving a pivotal role in our robustness benchmarking.

\subsubsection{Evaluation Settings} 

In our research, we select two self-supervised monocular depth estimation methods to assess their robustness under challenging conditions: the widely recognized MonoDepth2~\cite{godard2019digging}, and the AF-SfMLearner~\cite{shao2022self}, which is specifically tailored for endoscopic scenes.
Our evaluation adheres to the protocols of AF-SfMLearner~\cite{shao2022self}, ensuring the settings are directly comparable. We empirically set the accuracy weight $W_k$ in Eq.~\eqref{eq:acc} as $[0.5, 0.3, 0.2]$ respectively, and the robustness factor $\lambda$ as $1$ in Eq.~\eqref{eq:rob}. The experiments are conducted with an RTX4090 GPU on Ubuntu 22.04 OS.

\begin{figure}[t]
    \centering
    \includegraphics[width=\linewidth]{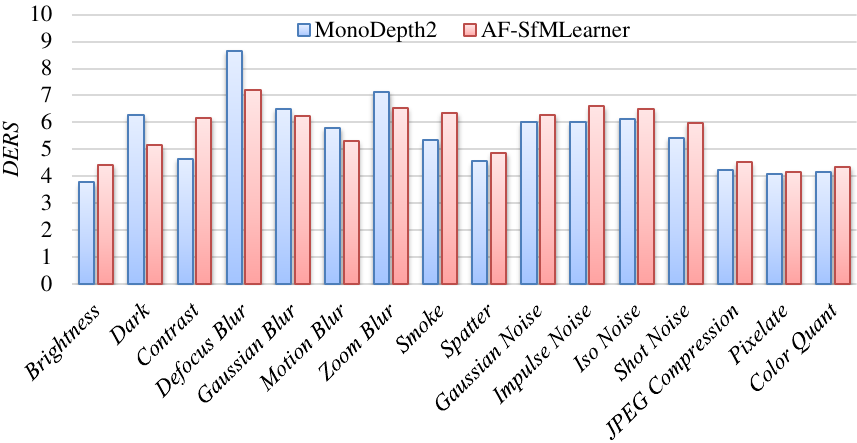}
    \caption{Robustness comparison of MonoDepth2~\cite{godard2019digging} and AF-SfMLearner~\cite{shao2022self} with the proposed \textit{DERS} metric.}
    \label{fig:ders}
\end{figure}

\section{Results and Analysis}
We evaluate the \textit{DERS} metric using MonoDepth2~\cite{godard2019digging} and AF-SfMLearner~\cite{shao2022self} across 16 types of corruptions, with the results depicted in Fig.~\ref{fig:ders}. Note that a smaller \textit{DERS} corresponds to more robust performance. The results reveal that endoscopic depth estimation models exhibit varying sensitivities to different corruptions. In other words, the robustness of a model against one type of corruption does not guarantee similar performance across other types. Generally, \textit{Blur} and \textit{Noise}, representing optical distortions and sensor \& electronic noise, significantly impact both model performance. Conversely, Data Compression and Digital Artifacts, such as \textit{JPEG Compression}, \textit{Pixelated}, and \textit{Color Quantization}, have the least effect on \textit{DERS}, suggesting minimal relevance in detracting from model robustness. 

For all 16 types of corruption, MonoDepth2~\cite{godard2019digging} outperform AF-SfMLearner~\cite{shao2022self} for most of the data corruptions, except for the \textit{Drak} and four \textit{Blur} corruption. 
Concretely, by averaging the \textit{DERS} scores across all corruption types, we find that MonoDepth2~\cite{godard2019digging} behaves more robustly and scores a lower \textit{DERS} of 5.55, compared to AF-SfMLearner’s 5.66. This indicates that models specifically designed and trained on uncorrupted data may overfit, whereas models with a generalist approach exhibit increased robustness to synthetic data shifts.

Additionally, we examine the impact of Brightness corruption on the MonoDepth2~\cite{godard2019digging} model. As indicated in Table~\ref{tab:min_corr}, it is noteworthy that the model performs better under specific levels of corruption than with unaltered data. This suggests that the model may prefer conditions with increased brightness. This finding also hints at the potential utility of implementing certain data perturbations during test-time inference to improve model performance. 
We have provided detailed results of both models for all data corruptions in the Appendix~\ref{sec:app}.

\begin{table}[t]
  \centering
  \caption{Performance of MonoDepth2~\cite{godard2019digging} under various levels of \textit{Brightness} corruption.}
  \resizebox{0.7\textwidth}{!}{%
    \begin{tabular}{c|cccc|ccc}
    \toprule
    Severity & AbsRel $\downarrow$ & SqRel $\downarrow$ & RMSE $\downarrow$ & LogRMSE $\downarrow$ & a1  $\uparrow$  & a2 $\uparrow$   & a3 $\uparrow$ \\
    \hline
    0     & 0.069 & 0.584 & 5.574 & 0.094 & 0.947 & 0.998 & 1.000 \\
    \hline
    1     & 0.064 & 0.498 & 5.182 & 0.088 & 0.957 & 0.998 & 1.000 \\
    2     & 0.063 & 0.507 & 5.291 & 0.088 & 0.960 & 0.996 & 1.000 \\
    3     & 0.065 & 0.571 & 5.655 & 0.093 & 0.958 & 0.994 & 0.999 \\
    4     & 0.068 & 0.638 & 6.014 & 0.098 & 0.953 & 0.993 & 0.998 \\
    5     & 0.070 & 0.699 & 6.330 & 0.102 & 0.949 & 0.992 & 0.997 \\
    \bottomrule
    \end{tabular}%
    }
  \label{tab:min_corr}%
\end{table}%

\section{Conclusion}

In summary, our investigation advances the field of endoscopic surgery by benchmarking the resilience of monocular depth estimation models against synthetic image corruption based on a widely-used endoscopic dataset, i.e., SCARED~\cite{allan2021stereo}. The proposed Depth Estimation Robustness Score (\textit{DERS}) serves as an enhanced evaluation metric that integrates error, accuracy, and robustness. 
Through rigorous analysis, we have exposed potential vulnerabilities and resilience in existing models under distortive conditions.
The study's scope, limited by the synthetic nature of the dataset's corruptions, invites future refinement. Real intraoperative data could further enrich the authenticity of our findings, while elaboration on the \textit{DERS} could offer more granular clinical insights. 

Robustness is essential and demanding during model deployment in real-world medical sites.
Looking ahead, we aim to broaden the dataset with more realistic corruptions and finetune the robustness metric to reflect advances in endoscopic technology and machine learning. By doing so, we strive to bridge the gap between computational robustness and clinical efficacy, ensuring depth estimation models contribute substantively to surgical precision and patient safety.

\begin{credits}
\subsubsection{\ackname} This work was supported by Hong Kong RGC CRF C4026-21G, GRF 14211420 \& 14203323; Shenzhen-Hong Kong-Macau Technology Research Programme (Type C) STIC Grant SGDX20210823103535014 (202108233000303).
\subsubsection{\discintname}
The authors have no competing interests to declare.
\end{credits}

\newpage

\appendix
\section{Appendix}\label{sec:app}
\begin{table}[h]
  \centering
  \caption{Quantitative results of MonoDepth2. Severity 0 corresponds to results on uncorrupted data. \textit{DERS} are reported following respective data corruption.}
  \resizebox{0.89\textwidth}{!}{%
    \begin{tabular}{c|cccc|ccc|cccc|ccc}
    \hline
    Severity & AbsRel $\downarrow$ & SqRel $\downarrow$ & RMSE $\downarrow$ & LogRMSE $\downarrow$ & a1  $\uparrow$  & a2 $\uparrow$   & a3 $\uparrow$    & AbsRel $\downarrow$ & SqRel $\downarrow$ & RMSE $\downarrow$ & LogRMSE $\downarrow$ & a1  $\uparrow$  & a2 $\uparrow$   & a3 $\uparrow$ \\
    \hline
          & \multicolumn{7}{c|}{\textit{Brightness} - 3.78}               & \multicolumn{7}{c}{\textit{Contrast} - 4.63} \\
    \hline
    0     & 0.069 & 0.584 & 5.574 & 0.094 & 0.947 & 0.998 & 1     & 0.069 & 0.584 & 5.574 & 0.094 & 0.947 & 0.998 & 1 \\
    1     & 0.064 & 0.498 & 5.182 & 0.088 & 0.957 & 0.998 & 1.000 & 0.065 & 0.594 & 5.774 & 0.095 & 0.955 & 0.994 & 0.998 \\
    2     & 0.063 & 0.507 & 5.291 & 0.088 & 0.960 & 0.996 & 1.000 & 0.073 & 0.725 & 6.423 & 0.105 & 0.946 & 0.992 & 0.997 \\
    3     & 0.065 & 0.571 & 5.655 & 0.093 & 0.958 & 0.994 & 0.999 & 0.087 & 0.986 & 7.516 & 0.123 & 0.922 & 0.990 & 0.996 \\
    4     & 0.068 & 0.638 & 6.014 & 0.098 & 0.953 & 0.993 & 0.998 & 0.099 & 1.237 & 8.412 & 0.137 & 0.898 & 0.985 & 0.995 \\
    5     & 0.070 & 0.699 & 6.330 & 0.102 & 0.949 & 0.992 & 0.997 & 0.107 & 1.424 & 9.064 & 0.148 & 0.881 & 0.979 & 0.995 \\
    \hline
          & \multicolumn{7}{c|}{\textit{Dark} - 6.29}                              & \multicolumn{7}{c}{\textit{Defocus Blur} - 8.64} \\
    \hline
    0     & 0.069 & 0.584 & 5.574 & 0.094 & 0.947 & 0.998 & 1     & 0.069 & 0.584 & 5.574 & 0.094 & 0.947 & 0.998 & 1 \\
    1     & 0.092 & 1.037 & 7.646 & 0.126 & 0.909 & 0.989 & 0.997 & 0.133 & 2.043 & 10.236 & 0.158 & 0.823 & 0.985 & 0.999 \\
    2     & 0.097 & 1.159 & 8.133 & 0.135 & 0.896 & 0.987 & 0.996 & 0.146 & 2.608 & 11.342 & 0.171 & 0.794 & 0.973 & 0.997 \\
    3     & 0.105 & 1.348 & 8.791 & 0.146 & 0.878 & 0.982 & 0.996 & 0.168 & 3.465 & 13.138 & 0.196 & 0.753 & 0.950 & 0.994 \\
    4     & 0.099 & 1.276 & 8.472 & 0.137 & 0.895 & 0.982 & 0.995 & 0.192 & 4.652 & 15.256 & 0.222 & 0.719 & 0.925 & 0.985 \\
    5     & 0.095 & 1.217 & 8.301 & 0.134 & 0.901 & 0.981 & 0.995 & 0.212 & 5.694 & 16.852 & 0.242 & 0.689 & 0.907 & 0.975 \\
    \hline
          & \multicolumn{7}{c|}{\textit{Gaussian Blur} - 6.49}                    & \multicolumn{7}{c}{\textit{Motion Blur} - 5.79} \\
    \hline
    0     & 0.069 & 0.584 & 5.574 & 0.094 & 0.947 & 0.998 & 1     & 0.069 & 0.584 & 5.574 & 0.094 & 0.947 & 0.998 & 1 \\
    1     & 0.096 & 0.983 & 7.445 & 0.127 & 0.905 & 0.994 & 1.000 & 0.080 & 0.736 & 6.310 & 0.110 & 0.929 & 0.993 & 1.000 \\
    2     & 0.142 & 2.462 & 11.056 & 0.168 & 0.803 & 0.976 & 0.998 & 0.099 & 1.063 & 7.546 & 0.133 & 0.886 & 0.987 & 0.999 \\
    3     & 0.162 & 3.288 & 12.764 & 0.190 & 0.763 & 0.955 & 0.995 & 0.123 & 1.571 & 9.099 & 0.160 & 0.829 & 0.975 & 0.998 \\
    4     & 0.183 & 4.242 & 14.586 & 0.213 & 0.731 & 0.933 & 0.988 & 0.141 & 2.062 & 10.367 & 0.179 & 0.794 & 0.963 & 0.997 \\
    5     & 0.226 & 6.385 & 17.830 & 0.256 & 0.663 & 0.895 & 0.969 & 0.149 & 2.285 & 10.987 & 0.188 & 0.770 & 0.960 & 0.997 \\
    \hline
          & \multicolumn{7}{c|}{\textit{Zoom Blur} - 7.13}                        & \multicolumn{7}{c}{\textit{Smoke} - 5.33} \\
    \hline
    0     & 0.069 & 0.584 & 5.574 & 0.094 & 0.947 & 0.998 & 1     & 0.069 & 0.584 & 5.574 & 0.094 & 0.947 & 0.998 & 1 \\
    1     & 0.106 & 1.180 & 7.976 & 0.142 & 0.877 & 0.981 & 0.999 & 0.078 & 0.768 & 6.506 & 0.108 & 0.938 & 0.993 & 0.998 \\
    2     & 0.116 & 1.396 & 8.668 & 0.152 & 0.856 & 0.978 & 0.999 & 0.085 & 0.916 & 7.064 & 0.118 & 0.924 & 0.990 & 0.997 \\
    3     & 0.111 & 1.288 & 8.299 & 0.150 & 0.861 & 0.974 & 0.999 & 0.095 & 1.120 & 7.829 & 0.132 & 0.901 & 0.986 & 0.996 \\
    4     & 0.117 & 1.421 & 8.727 & 0.157 & 0.847 & 0.971 & 0.999 & 0.096 & 1.124 & 7.876 & 0.133 & 0.900 & 0.985 & 0.996 \\
    5     & 0.113 & 1.317 & 8.361 & 0.153 & 0.856 & 0.971 & 0.999 & 0.104 & 1.319 & 8.644 & 0.146 & 0.882 & 0.980 & 0.994 \\
    \hline
          & \multicolumn{7}{c|}{\textit{Spatter} - 4.55}                           & \multicolumn{7}{c}{\textit{Gaussian Noise} - 6.01} \\
    \hline
    0     & 0.069 & 0.584 & 5.574 & 0.094 & 0.947 & 0.998 & 1     & 0.069 & 0.584 & 5.574 & 0.094 & 0.947 & 0.998 & 1 \\
    1     & 0.068 & 0.563 & 5.504 & 0.093 & 0.952 & 0.997 & 1.000 & 0.084 & 0.970 & 7.525 & 0.123 & 0.924 & 0.987 & 0.996 \\
    2     & 0.074 & 0.828 & 6.848 & 0.109 & 0.938 & 0.990 & 0.996 & 0.090 & 1.102 & 7.946 & 0.128 & 0.913 & 0.984 & 0.995 \\
    3     & 0.084 & 0.986 & 7.533 & 0.120 & 0.925 & 0.987 & 0.996 & 0.093 & 1.144 & 8.031 & 0.129 & 0.911 & 0.984 & 0.996 \\
    4     & 0.081 & 0.958 & 7.436 & 0.118 & 0.932 & 0.987 & 0.996 & 0.093 & 1.136 & 7.983 & 0.128 & 0.911 & 0.985 & 0.996 \\
    5     & 0.082 & 0.979 & 7.518 & 0.119 & 0.931 & 0.986 & 0.996 & 0.093 & 1.129 & 7.944 & 0.128 & 0.910 & 0.985 & 0.996 \\
    \hline
          & \multicolumn{7}{c|}{\textit{Impulse Noise} - 6.03}                    & \multicolumn{7}{c}{\textit{Iso Noise} - 6.14} \\
    \hline
    0     & 0.069 & 0.584 & 5.574 & 0.094 & 0.947 & 0.998 & 1     & 0.069 & 0.584 & 5.574 & 0.094 & 0.947 & 0.998 & 1 \\
    1     & 0.084 & 0.988 & 7.608 & 0.122 & 0.920 & 0.988 & 0.996 & 0.090 & 1.029 & 7.710 & 0.125 & 0.919 & 0.989 & 0.996 \\
    2     & 0.091 & 1.074 & 7.912 & 0.128 & 0.914 & 0.985 & 0.996 & 0.092 & 1.078 & 7.851 & 0.127 & 0.916 & 0.987 & 0.996 \\
    3     & 0.093 & 1.112 & 7.972 & 0.128 & 0.912 & 0.985 & 0.996 & 0.093 & 1.128 & 7.985 & 0.128 & 0.911 & 0.985 & 0.996 \\
    4     & 0.094 & 1.110 & 7.911 & 0.127 & 0.913 & 0.986 & 0.996 & 0.094 & 1.141 & 7.997 & 0.128 & 0.911 & 0.985 & 0.996 \\
    5     & 0.093 & 1.105 & 7.868 & 0.126 & 0.912 & 0.986 & 0.996 & 0.093 & 1.135 & 7.965 & 0.128 & 0.910 & 0.985 & 0.996 \\
    \hline
          & \multicolumn{7}{c|}{\textit{Shot Noise} - 5.43}                       & \multicolumn{7}{c}{\textit{JPEG Compression} - 4.24} \\
    \hline
    0     & 0.069 & 0.584 & 5.574 & 0.094 & 0.947 & 0.998 & 1     & 0.069 & 0.584 & 5.574 & 0.094 & 0.947 & 0.998 & 1 \\
    1     & 0.082 & 0.792 & 6.736 & 0.114 & 0.937 & 0.992 & 0.999 & 0.067 & 0.573 & 5.634 & 0.094 & 0.952 & 0.996 & 0.999 \\
    2     & 0.088 & 0.935 & 7.371 & 0.121 & 0.926 & 0.992 & 0.997 & 0.069 & 0.635 & 6.013 & 0.099 & 0.949 & 0.995 & 0.998 \\
    3     & 0.092 & 1.049 & 7.740 & 0.125 & 0.918 & 0.989 & 0.996 & 0.071 & 0.697 & 6.340 & 0.103 & 0.944 & 0.993 & 0.998 \\
    4     & 0.092 & 1.099 & 7.855 & 0.126 & 0.914 & 0.986 & 0.996 & 0.078 & 0.829 & 6.997 & 0.114 & 0.935 & 0.990 & 0.997 \\
    5     & 0.092 & 1.110 & 7.881 & 0.126 & 0.912 & 0.985 & 0.996 & 0.087 & 0.964 & 7.584 & 0.124 & 0.922 & 0.989 & 0.996 \\
    \hline
          & \multicolumn{7}{c|}{\textit{Pixelate} - 4.07}                          & \multicolumn{7}{c}{\textit{Color Quant} - 4.17} \\
    \hline
    0     & 0.069 & 0.584 & 5.574 & 0.094 & 0.947 & 0.998 & 1     & 0.069 & 0.584 & 5.574 & 0.094 & 0.947 & 0.998 & 1 \\
    1     & 0.069 & 0.564 & 5.499 & 0.094 & 0.952 & 0.998 & 1.000 & 0.071 & 0.601 & 5.663 & 0.096 & 0.946 & 0.998 & 1.000 \\
    2     & 0.069 & 0.564 & 5.523 & 0.094 & 0.953 & 0.997 & 1.000 & 0.073 & 0.640 & 5.869 & 0.099 & 0.942 & 0.997 & 1.000 \\
    3     & 0.071 & 0.585 & 5.654 & 0.095 & 0.952 & 0.997 & 1.000 & 0.073 & 0.638 & 5.955 & 0.099 & 0.947 & 0.996 & 0.999 \\
    4     & 0.075 & 0.637 & 5.947 & 0.100 & 0.948 & 0.997 & 1.000 & 0.074 & 0.711 & 6.348 & 0.102 & 0.954 & 0.993 & 0.998 \\
    5     & 0.079 & 0.697 & 6.248 & 0.105 & 0.943 & 0.996 & 0.999 & 0.091 & 0.988 & 7.548 & 0.123 & 0.927 & 0.991 & 0.997 \\
    \hline
    \end{tabular}%
    }
  \label{tab:res_md2}%
\end{table}%

\begin{table}[htbp]
  \centering
    \caption{Quantitative results of AF-SfMLearner. Severity 0 corresponds to results on uncorrupted data. \textit{DERS} are reported following respective data corruption.}
  \resizebox{\textwidth}{!}{%
    \begin{tabular}{c|cccc|ccc|cccc|ccc}
    \hline
    Severity & AbsRel $\downarrow$ & SqRel $\downarrow$ & RMSE $\downarrow$ & LogRMSE $\downarrow$ & a1  $\uparrow$  & a2 $\uparrow$   & a3 $\uparrow$    & AbsRel $\downarrow$ & SqRel $\downarrow$ & RMSE $\downarrow$ & LogRMSE $\downarrow$ & a1  $\uparrow$  & a2 $\uparrow$   & a3 $\uparrow$ \\
    \hline
          & \multicolumn{7}{c|}{Brightness - 4.42}                 & \multicolumn{7}{c}{Contrast - 6.17} \\
    \hline
    0     & 0.059 & 0.448 & 5.012 & 0.082 & 0.973 & 0.997 & 0.999 & 0.059 & 0.448 & 5.012 & 0.082 & 0.973 & 0.997 & 0.999 \\
    1     & 0.059 & 0.465 & 5.132 & 0.083 & 0.972 & 0.996 & 0.999 & 0.062 & 0.585 & 5.809 & 0.091 & 0.964 & 0.994 & 0.998 \\
    2     & 0.061 & 0.521 & 5.451 & 0.087 & 0.969 & 0.995 & 0.999 & 0.066 & 0.660 & 6.182 & 0.097 & 0.956 & 0.993 & 0.998 \\
    3     & 0.064 & 0.591 & 5.835 & 0.093 & 0.965 & 0.994 & 0.998 & 0.075 & 0.810 & 6.840 & 0.108 & 0.940 & 0.991 & 0.997 \\
    4     & 0.069 & 0.674 & 6.274 & 0.100 & 0.959 & 0.992 & 0.997 & 0.094 & 1.187 & 8.208 & 0.133 & 0.900 & 0.983 & 0.996 \\
    5     & 0.074 & 0.756 & 6.661 & 0.107 & 0.952 & 0.991 & 0.997 & 0.114 & 1.622 & 9.469 & 0.152 & 0.853 & 0.977 & 0.994 \\
    \hline
          & \multicolumn{7}{c|}{Dark - 5.16}                       & \multicolumn{7}{c}{Defocus Blur - 7.20} \\
    \hline
    0     & 0.059 & 0.448 & 5.012 & 0.082 & 0.973 & 0.997 & 0.999 & 0.059 & 0.448 & 5.012 & 0.082 & 0.973 & 0.997 & 0.999 \\
    1     & 0.073 & 0.747 & 6.59  & 0.105 & 0.953 & 0.991 & 0.997 & 0.084 & 0.757 & 6.61  & 0.115 & 0.922 & 0.994 & 1 \\
    2     & 0.078 & 0.861 & 7.109 & 0.113 & 0.941 & 0.989 & 0.996 & 0.103 & 1.053 & 7.713 & 0.138 & 0.879 & 0.989 & 1.000 \\
    3     & 0.088 & 1.065 & 7.897 & 0.127 & 0.921 & 0.983 & 0.996 & 0.127 & 1.571 & 9.270 & 0.167 & 0.812 & 0.975 & 0.999 \\
    4     & 0.099 & 1.316 & 8.657 & 0.140 & 0.890 & 0.979 & 0.994 & 0.144 & 2.006 & 10.399 & 0.189 & 0.769 & 0.954 & 0.998 \\
    5     & 0.101 & 1.339 & 8.658 & 0.139 & 0.882 & 0.980 & 0.995 & 0.156 & 2.321 & 11.185 & 0.204 & 0.738 & 0.942 & 0.997 \\
    \hline
          & \multicolumn{7}{c|}{Gaussian Blur - 6.25}              & \multicolumn{7}{c}{Motion Blur - 6.31} \\
    \hline
    0     & 0.059 & 0.448 & 5.012 & 0.082 & 0.973 & 0.997 & 0.999 & 0.059 & 0.448 & 5.012 & 0.082 & 0.973 & 0.997 & 0.999 \\
    1     & 0.064 & 0.513 & 5.451 & 0.091 & 0.958 & 0.997 & 0.999 & 0.064 & 0.515 & 5.391 & 0.09  & 0.959 & 0.997 & 0.999 \\
    2     & 0.096 & 0.937 & 7.302 & 0.130 & 0.895 & 0.992 & 1.000 & 0.073 & 0.635 & 5.979 & 0.103 & 0.942 & 0.995 & 1.000 \\
    3     & 0.123 & 1.466 & 8.976 & 0.162 & 0.825 & 0.978 & 1.000 & 0.088 & 0.864 & 6.938 & 0.122 & 0.915 & 0.988 & 0.999 \\
    4     & 0.142 & 1.954 & 10.269 & 0.186 & 0.772 & 0.957 & 0.999 & 0.105 & 1.169 & 8.092 & 0.145 & 0.878 & 0.979 & 0.999 \\
    5     & 0.166 & 2.567 & 11.717 & 0.213 & 0.708 & 0.935 & 0.997 & 0.114 & 1.383 & 8.770 & 0.157 & 0.853 & 0.972 & 0.998 \\
    \hline
          & \multicolumn{7}{c|}{Zoom Blur - 6.53}                  & \multicolumn{7}{c}{Smoke - 6.35} \\
    \hline
    0     & 0.059 & 0.448 & 5.012 & 0.082 & 0.973 & 0.997 & 0.999 & 0.059 & 0.448 & 5.012 & 0.082 & 0.973 & 0.997 & 0.999 \\
    1     & 0.082 & 0.781 & 6.682 & 0.119 & 0.920 & 0.988 & 0.999 & 0.077 & 0.775 & 6.562 & 0.107 & 0.940 & 0.992 & 0.998 \\
    2     & 0.089 & 0.917 & 7.229 & 0.131 & 0.904 & 0.982 & 0.998 & 0.084 & 0.921 & 7.128 & 0.118 & 0.925 & 0.990 & 0.997 \\
    3     & 0.089 & 0.906 & 7.184 & 0.130 & 0.908 & 0.980 & 0.999 & 0.093 & 1.081 & 7.716 & 0.129 & 0.906 & 0.986 & 0.996 \\
    4     & 0.094 & 1.001 & 7.544 & 0.138 & 0.898 & 0.976 & 0.998 & 0.093 & 1.096 & 7.778 & 0.129 & 0.904 & 0.987 & 0.997 \\
    5     & 0.093 & 0.949 & 7.360 & 0.134 & 0.906 & 0.979 & 0.998 & 0.099 & 1.193 & 8.181 & 0.136 & 0.894 & 0.984 & 0.996 \\
    \hline
          & \multicolumn{7}{c|}{Spatter - 4.87}                    & \multicolumn{7}{c}{Gaussian Noise - 6.29} \\
    \hline
    0     & 0.059 & 0.448 & 5.012 & 0.082 & 0.973 & 0.997 & 0.999 & 0.059 & 0.448 & 5.012 & 0.082 & 0.973 & 0.997 & 0.999 \\
    1     & 0.059 & 0.472 & 5.187 & 0.084 & 0.973 & 0.996 & 0.999 & 0.077 & 0.793 & 6.799 & 0.110 & 0.946 & 0.992 & 0.997 \\
    2     & 0.067 & 0.698 & 6.313 & 0.099 & 0.959 & 0.991 & 0.996 & 0.082 & 0.900 & 7.268 & 0.117 & 0.937 & 0.988 & 0.996 \\
    3     & 0.073 & 0.832 & 6.881 & 0.108 & 0.947 & 0.989 & 0.996 & 0.085 & 0.991 & 7.583 & 0.122 & 0.929 & 0.986 & 0.996 \\
    4     & 0.075 & 0.848 & 7.018 & 0.110 & 0.946 & 0.989 & 0.996 & 0.09  & 1.077 & 7.873 & 0.127 & 0.921 & 0.985 & 0.995 \\
    5     & 0.076 & 0.857 & 7.042 & 0.111 & 0.945 & 0.989 & 0.996 & 0.095 & 1.192 & 8.240 & 0.133 & 0.910 & 0.983 & 0.995 \\
    \hline
          & \multicolumn{7}{c|}{Impulse Noise - 6.61}              & \multicolumn{7}{c}{Iso Noise - 6.49} \\
    \hline
    0     & 0.059 & 0.448 & 5.012 & 0.082 & 0.973 & 0.997 & 0.999 & 0.059 & 0.448 & 5.012 & 0.082 & 0.973 & 0.997 & 0.999 \\
    1     & 0.078 & 0.877 & 7.127 & 0.113 & 0.939 & 0.989 & 0.996 & 0.082 & 0.875 & 7.196 & 0.116 & 0.939 & 0.991 & 0.997 \\
    2     & 0.084 & 0.976 & 7.523 & 0.120 & 0.931 & 0.986 & 0.996 & 0.084 & 0.914 & 7.341 & 0.118 & 0.937 & 0.989 & 0.996 \\
    3     & 0.087 & 1.027 & 7.706 & 0.123 & 0.925 & 0.985 & 0.996 & 0.085 & 0.957 & 7.486 & 0.120 & 0.932 & 0.987 & 0.996 \\
    4     & 0.091 & 1.099 & 7.952 & 0.128 & 0.918 & 0.985 & 0.995 & 0.088 & 1.018 & 7.676 & 0.123 & 0.926 & 0.986 & 0.996 \\
    5     & 0.093 & 1.157 & 8.128 & 0.131 & 0.912 & 0.984 & 0.995 & 0.092 & 1.116 & 8.016 & 0.129 & 0.916 & 0.984 & 0.995 \\
    \hline
          & \multicolumn{7}{c|}{Shot Noise - 5.98}                 & \multicolumn{7}{c}{JPEG Compression - 4.51} \\
    \hline
    0     & 0.059 & 0.448 & 5.012 & 0.082 & 0.973 & 0.997 & 0.999 & 0.059 & 0.448 & 5.012 & 0.082 & 0.973 & 0.997 & 0.999 \\
    1     & 0.075 & 0.710 & 6.462 & 0.104 & 0.954 & 0.994 & 0.998 & 0.059 & 0.513 & 5.455 & 0.087 & 0.970 & 0.995 & 0.998 \\
    2     & 0.081 & 0.835 & 7.044 & 0.113 & 0.943 & 0.992 & 0.997 & 0.06  & 0.555 & 5.665 & 0.089 & 0.966 & 0.994 & 0.998 \\
    3     & 0.086 & 0.928 & 7.393 & 0.119 & 0.934 & 0.989 & 0.997 & 0.061 & 0.580 & 5.800 & 0.091 & 0.964 & 0.994 & 0.998 \\
    4     & 0.09  & 1.022 & 7.711 & 0.125 & 0.923 & 0.987 & 0.996 & 0.065 & 0.629 & 6.078 & 0.095 & 0.959 & 0.993 & 0.998 \\
    5     & 0.094 & 1.130 & 8.067 & 0.131 & 0.914 & 0.984 & 0.996 & 0.073 & 0.748 & 6.619 & 0.104 & 0.946 & 0.993 & 0.998 \\
    \hline
          & \multicolumn{7}{c|}{Pixelate - 4.16}                   & \multicolumn{7}{c}{Color Quant - 4.33} \\
    \hline
    0     & 0.059 & 0.448 & 5.012 & 0.082 & 0.973 & 0.997 & 0.999 & 0.059 & 0.448 & 5.012 & 0.082 & 0.973 & 0.997 & 0.999 \\
    1     & 0.059 & 0.453 & 5.099 & 0.083 & 0.973 & 0.997 & 0.999 & 0.059 & 0.464 & 5.109 & 0.084 & 0.972 & 0.997 & 0.999 \\
    2     & 0.059 & 0.457 & 5.133 & 0.084 & 0.972 & 0.997 & 0.999 & 0.06  & 0.495 & 5.293 & 0.086 & 0.969 & 0.996 & 0.999 \\
    3     & 0.061 & 0.481 & 5.314 & 0.086 & 0.970 & 0.997 & 0.999 & 0.063 & 0.547 & 5.604 & 0.090 & 0.968 & 0.995 & 0.999 \\
    4     & 0.063 & 0.518 & 5.539 & 0.089 & 0.966 & 0.996 & 0.999 & 0.074 & 0.811 & 6.812 & 0.108 & 0.944 & 0.990 & 0.996 \\
    5     & 0.063 & 0.515 & 5.506 & 0.088 & 0.965 & 0.996 & 0.999 & 0.088 & 1.059 & 7.770 & 0.124 & 0.921 & 0.987 & 0.996 \\
    \hline
    \end{tabular}%
    }
  \label{tab:res_afs}%
\end{table}%

\newpage

\bibliography{mybib}{}
\bibliographystyle{splncs04}

\end{document}